\documentclass[10pt,twocolumn,letterpaper]{article}

\usepackage{wacv}
\usepackage{times}
\usepackage{epsfig}
\usepackage{graphicx}
\usepackage{amsmath}
\usepackage{amssymb}
\usepackage{booktabs}

\usepackage[noadjust]{cite}
\usepackage{algorithm}
\usepackage{algpseudocode}
\usepackage{tabularx}
\usepackage{caption} 
\usepackage{mwe}%
\usepackage[accsupp]{axessibility}

\def\wacvPaperID{2} 

\wacvfinalcopy 

\ifwacvfinal
\usepackage[breaklinks=true,bookmarks=false]{hyperref}
\else
\usepackage[pagebackref=true,breaklinks=true,colorlinks,bookmarks=false]{hyperref}
\fi

\begin{document}

\title{Improving the Detection of Small Oriented Objects in Aerial Images}

\author{Chandler Timm C. Doloriel and Rhandley D. Cajote\\
Electrical and Electronics Engineering Institute\\
University of the Philippines, Diliman, Philippines\\
{\tt\small \{chandler.timm.doloriel, rhandley.cajote\}@eee.upd.edu.ph}
}

\maketitle
\thispagestyle{empty}

\begin{abstract}
    Small oriented objects that represent tiny pixel-area in large-scale aerial images are difficult to detect due to their size and orientation. Existing oriented aerial detectors have shown promising results but are mainly focused on orientation modeling with less regard to the size of the objects. In this work, we proposed a method to accurately detect small oriented objects in aerial images by enhancing the classification and regression tasks of the oriented object detection model. We designed the Attention-Points Network consisting of two losses: Guided-Attention Loss (GALoss) and Box-Points Loss (BPLoss). GALoss uses an instance segmentation mask as ground-truth to learn the attention features needed to improve the detection of small objects. These attention features are then used to predict box points for BPLoss, which determines the points' position relative to the target oriented bounding box. Experimental results show the effectiveness of our Attention-Points Network on a standard oriented aerial dataset with small object instances (DOTA-v1.5) and on a maritime-related dataset (HRSC2016). The code is publicly available\footnote {https://github.com/chandlerbing65nm/APDetection.}.
\end{abstract}

\section{Introduction}

Object detection is a valuable technique for understanding objects in an image, describing both what and where these objects are with the goal of identifying the location enclosed in bounding boxes. The usual method is to use a rectangle bounding box with no angle orientation, also called a horizontal bounding box (HBB). To enclose an object inside an HBB, the model should be able to accurately locate the object and identify its class. There are many use-cases of using HBB, specific examples are in applications for vehicle tracking\cite{Chen_2021_CVPR, Wu_2021_CVPR}, face recognition\cite{Liu_2022_CVPR, HLAFace_2021_CVPR}, maritime detection \cite{Zust_2022_WACV, Varga_2022_WACV} etc., even the objects in aerial images can be detected using an HBB. However, it is very ineffective to detect oriented aerial objects using this method, objects cannot be precisely localized, more noise and background will be enclosed that can lead to misdetection. Therefore, an object detector that can produce an oriented bounding box (OBB) is needed to detect oriented aerial objects.

Oriented aerial detection is a popular research topic in computer vision in the past years \cite{Xia2018,Han2021a,Ma2018,Xu2021,Yang2021a,Ding2019,Yi2020,Xie2021,Li2021,Chen2020,Yang2021b,Han2021b,Yang2021c,Yang2019}. Existing methods have designed effective OBB detectors that can accurately enclose oriented objects. These methods vary from refining features \cite{Yang2019, Han2021a, Yang2021b}, proposal extraction \cite{Ma2018, Ding2019,Han2021b,Li2021}, orientation alignment \cite{Xu2021,Li2021}, and regression loss design \cite{Yang2021a,Chen2020,Yang2021c}. However, despite being very effective in detecting oriented objects, more research is needed to detect small oriented objects in aerial images.



Objects in aerial images vary greatly in size, orientation, or surroundings. Existing methods have used DOTA-v1.0 \cite{Xia2018} to benchmark their performance in oriented object detection. However, DOTA-v1.0 is not known to contain small and complex instances. To properly benchmark our method in dataset with small oriented objects, we used DOTA-v1.5 \cite{Xia2018}.

In this work, we proposed the Attention-Points Network to detect small oriented objects in aerial images. We used attention mechanism to gather the important features of an object, which increases the model's awareness especially on hard-to-identify objects such as small and complex instances. Furthermore, the attention features are used in our designed regression loss to predict box points and score them based on their relative position to the target OBB.

In Attention-Points Network, we designed two loss functions: Guided-Attention Loss (GALoss) and Box-Points Loss (BPLoss). GALoss compares the attention features to target features that can be obtained using the instance segmentation masks. However, these masks are not easy to annotate due to the irregular shapes of aerial objects. Instead, we used coarse-level masks that only need the bounding box coordinates for annotation. Meanwhile, BPLoss is calculated by scoring the box points based on their relative position to the target OBB. We measure the relative position of the box points to the target OBB using a kernel derived from the sigmoid function and compute for the IoU-based loss.

To verify our work, we conducted experiments on the standard oriented aerial dataset, DOTA \cite{Xia2018}. We chose the version of this dataset that contains very small instances (less than 10 pixels), DOTA-v1.5. We also used Oriented RCNN \cite{Xie2021} as the baseline, and R-50-FPN \cite{he2016residual} as backbone. Results show the effectiveness of our Attention-Points Network on a standard oriented aerial dataset with small object instances (DOTA-v1.5) and on a maritime-related dataset (HRSC2016).

The contributions of this paper are summarized as follows:
\begin{enumerate}
    \item We proposed the Attention-Points Network to improve the detection of small oriented objects in aerial images. This network uses two losses: Guided-Attention Loss (GALoss) and Box-Points Loss (BPLoss). GALoss uses attention features to improve the detection of small objects and BPLoss is used to score the predicted box-points based on their relative position to the target OBB.
    \item We compared our method to other existing OBB detectors. Experimental results show the effectiveness of our Attention-Points Network on a standard oriented aerial dataset with small object instances (DOTA-v1.5) and on a maritime-related dataset (HRSC2016).
    \item We conducted an ablation experiment to evaluate our designed loss functions, GALoss and BPLoss, and compared them to the baseline architecture. Our results showed that each loss function contributes to the overall performance without lagging behind the baseline. 
\end{enumerate}

\section{Related Work}

In this section, we discuss the different approaches to detecting objects using a bounding box along with the following classification: generating proposal and regression loss design. With these, we further describe the methods in oriented object detection and lastly, we look at the details of attention mechanism and its use for oriented object detection in aerial images.

\subsection{Generating Region Proposals}

Generating region proposal uses an additional network to predict the location and class of objects. In \cite{Uijlings2013}, a segmentation map is applied to an image to discriminate the objects with the background and rejects the overlapping proposals with low objectness scores. Then, the intersection-over-union (IoU) between the ground-truth and prediction is computed with different thresholds. Usually, IoU$\ge$0.5 will be considered as an object class while IoU$<$0.5 is background. Finally, a convolutional neural network (CNN) is used to classify and localize the objects.

Current multi-stage methods tackle the issue of proposal generation. A region proposal network from \cite{Ren2017} is designed to share the convolution layers with the feature extractor to minimize the cost, that creates sets of RoIs by dictating the model where to look. It scans every location in the extracted features to assess whether further processing is needed in a certain region and uses \( k \) anchor boxes with two scores representing whether there's an object or not at each location.

\subsubsection{Oriented Proposals}

To represent the rotation of an object for detection is to use anchors that rely on an angle parameter \cite{Ma2018,Ding2019,Han2021b,Xie2021,Han2021a,Xu2021,Yang2021b}. Early method that generate proposals with fifty-four anchors with different scales, ratios, and angles \cite{Ma2018} obtained good performance in detecting objects that are arbitrary-oriented, but a  large number of anchors causes computational complexity and memory overhead. The transformation from horizontal to rotated RoI \cite{Ding2019} was seen as a solution to reduce the number of generated anchors since the angle parameter is not introduced in generating proposals. However, the transformation network is also heavy and complex because it involves fully connected layers and alignment operations during the learning of RoI's. Another approach based on the transformation network \cite{Ding2019} used a rotation equivariant feature extractor \cite{Han2021b} to draw out rotation-invariant features for region proposals. It warps and aligns the rotated region-of-interests in its correct orientation dimension through feature interpolation. However, it did not reduce the computational cost of the transformation network and the rotation equivariant backbone is computationally complex.

The key to computational bottleneck is the design of a more efficient architecture \cite{Xie2021}, this is the motivation to improve the previous oriented object detectors. To realize this, two-stage detection frameworks should generate high-quality proposals while quickly detecting objects in a cost-efficient manner \cite{Xie2021}. In this paper, we used the network from \cite{Xie2021} because of its efficient architecture in the proposal generation and further improve its detection using loss function design.

\subsection{Regression Loss Design}

A bounding box is predicted with a regression loss that gives the error between the ground-truth and prediction. Regression losses can be divided into two categories: L1-type and IoU-based loss. An example of an L1-type loss is the smooth L1 loss, also known as Huber loss, given as:

\begin{equation}
\label{eq:smoothl1}
loss(x,y) = 
\begin{cases}
0.5(x-y)^2, if |x-y|<1 \\
|x-y|-0.5, \text{otherwise}
\end{cases} 
\end{equation}

Smooth L1 loss is less sensitive to outliers and also prevents exploding gradients. If the absolute loss is greater than one, the loss function is not squared and avoids high-value losses hence preventing exploding gradients. However, this loss is uncorrelated with the metric used in object detection. A low loss value does not always correspond to a high metric. Thus, the IoU-based loss is designed to combine the regression loss and the metric, it is given as:

\begin{equation}
\label{eq:IoU_basic}
\mathcal{L}_{IoU} = 1 - IoU
\end{equation}

Using (\ref{eq:IoU_basic}) is simple but the IoU cannot be computed when there is no overlapping area between two bounding boxes and it cannot be used in oriented object detection because the resulting function becomes undifferentiable, meaning the gradients cannot be backpropagated to enable network training.

\subsubsection{Oriented IoU-based Loss}

There are three known IoU-based loss in oriented object detection in aerial images. The first calculates the polygon distance of the ground-truth and prediction. It partly circumvents the need for a differentiable IoU-based loss by combining with the smooth L1 loss \cite{Yang2019}. However, since the IoU-based loss is undifferentiable, the gradient direction is still dominated by the smooth L1 loss so the metric cannot be regarded as consistent. The second converts the ground-truth and prediction box into a 2-D Gaussian distribution and calculates the loss function through Wasserstein distance \cite{Yang2021c} and Kullback-Leibler divergence \cite{Yang2021a}. It approximates the resulting IoU-based loss to obtain a differentiable function so that it can produce useful gradients. The issue is complexity, the conversion of a bounding box to a gaussian distribution and the distance calculation using Wasserstein and Kullback-Leibler divergence are complicated and adds significant overhead in the network. The third calculates the IoU-based loss directly by accumulating the contribution of the overlapping pixels of the ground-truth and prediction box \cite{Chen2020}. The function used is the normal distance between the pixels and the OBB center which is simple to implement but it cannot accurately represent the target object’s importance since each pixel has the same level of attention. In this paper, we designed an IoU-based loss by predicting box-points from attention features. These box-points are then scored based on their relative position to the target OBB.

\subsection{Attention Mechanism}

Convolutional neural network (CNN) is a type of attention mechanism in computer vision that uses a filter to process the input features and calculates the non-linearity using an activation function. An example of object detection model that used CNN as attention mechanism is from \cite{Pang2019}, the work highlights the occluded objects that are detected by the region proposal network, improving the detection of occluded objects. 

The disadvantage of using CNN as an attention mechanism is that the filter sizes are limited, it can usually take 3x3 or 5x5 but the attention features become coarser when we increase the filter size. Thus, CNN's can only take the attention in the local space of a filter and farther features are ignored.

\subsubsection{Self-Attention}

Another type of attention mechanism is self-attention, used in natural language processing (NLP) \cite{Bahdanau2015} to solve the problem of recognizing long sentences. In machine translation, to predict the next word is to look at the previous words, but this can be a bottleneck if the sentence is too long because it will lose the information from the previous words. Thus, self-attention searches the whole sentence, both previous and succeeding words, and analyzes the context to predict the next word. It relates different positions of a word in a sentence in order to obtain richer information.

Self-attention has three elements: Queries (Q), Keys (K), and Values (V). At the start, the input sentence is transformed into a vector that represents the three elements, which calculates the attention scores that measures how much attention to put in a word from a certain location. To compute this, the Q and K of the word is multiplied using the dot product and normalized to make the gradients  stable. Then, the result is compared with V to highlight the words to focus and disregard irrelevant words. The steps above are formulated by (\ref{eq:selfamath}).

\begin{figure*}[t!]
\begin{center}
\includegraphics[width = .9\linewidth]{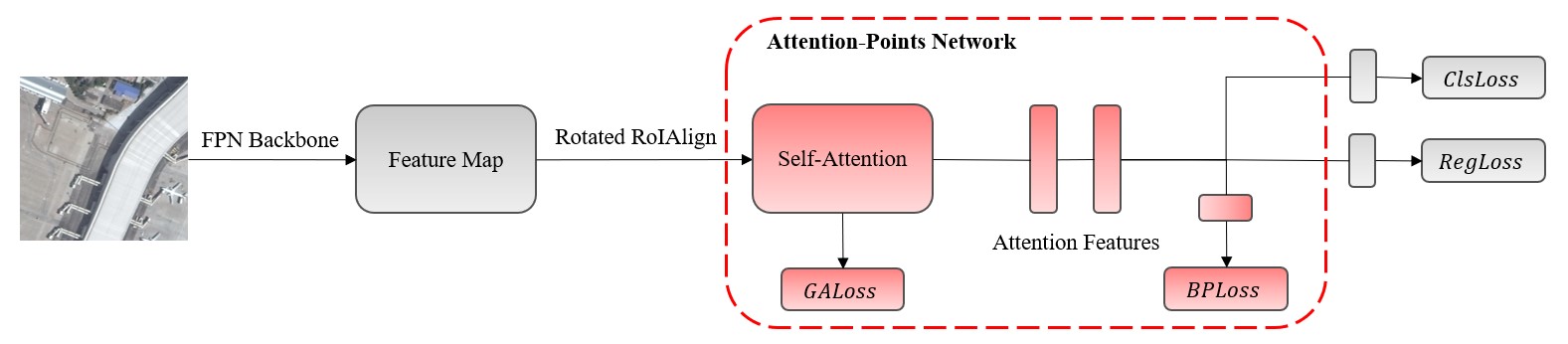}
\end{center}
\caption{Architecture of Attention-Points Network.}
\label{fig:attentionpointsnetworkjpg}
\end{figure*}

\begin{equation}
\label{eq:selfamath}
Attention(Q,K,V) = softmax(\frac{QK^T}{\sqrt{d_k}})V
\end{equation}

As self-attention had become effective in NLP, it also gained popularity in computer vision. In some cases, they completely replaced CNN's in image classification tasks. The architecture that used self-attention is the Vision Transformer (ViT) \cite{Zhai2021}. ViT transforms the input image into a series of patches and follows the same computation for Q, K, and V, then directly predicts the class label for the image. The self-attention in ViT makes it possible to embed information globally across an image.

The downside of self-attention is the computational cost because of its function to get the attention information globally, all features are used for computation compared to CNN that only used the local space. This can be mitigated by using small input image but that also limits the application of the model \cite{Wang2018}. To reduce computational cost, \cite{Zhuoran2021} swapped the positions of Q and V because the dot-product of K and V would result in a smaller dimension compared to using Q. It is equivalent to the self-attention but the dot-product of vectors is used differently.

We used the concept of efficient self-attention to gather the global features and detect small oriented objects in aerial images. Furthermore, we designed a loss function that can refine the attention features by comparing them to the segmentation masks of objects.
\section{Attention-Points Network}

We present the details of our proposed small oriented object detector with Attention-Points Network (APN) with two new loss functions: Guided-Attention Loss (GALoss) and Box-Points Loss (BPLoss). The baseline is from Oriented RCNN \cite{Xie2021} and we placed APN after the rotated RoIAlign, the architecture is shown in Fig. \ref{fig:attentionpointsnetworkjpg}. It is a two-stage detector consisting of feature extraction in the first stage and prediction in the second stage. We used ResNet \cite{he2016residual} as the backbone that produce five levels of features with each level going to the feature pyramid network (FPN) \cite{lin2016feature} for refined feature extraction. The features are inputs to the region proposal network (RPN) that generates proposals in various scales and ratios that tells the detector where the objects might be. These proposals are extracted and transformed into features by rotated RoIAlign operation and then used as input to the self-attention as region-of-interests (RoI).

We used feature size of 7x7 for each RoI to use as input to the self-attention and refine it using the \textit{\textbf{GALoss}}. The  attention features are used to predict box-points that are scored based on their relative position to the target OBB using \textit{\textbf{BPLoss}}.

\subsection{Guided-Attention Loss ( \textbf{\textit{GALoss}} )}

\begin{figure}[t!]
\centering
\includegraphics[width=0.9\linewidth]{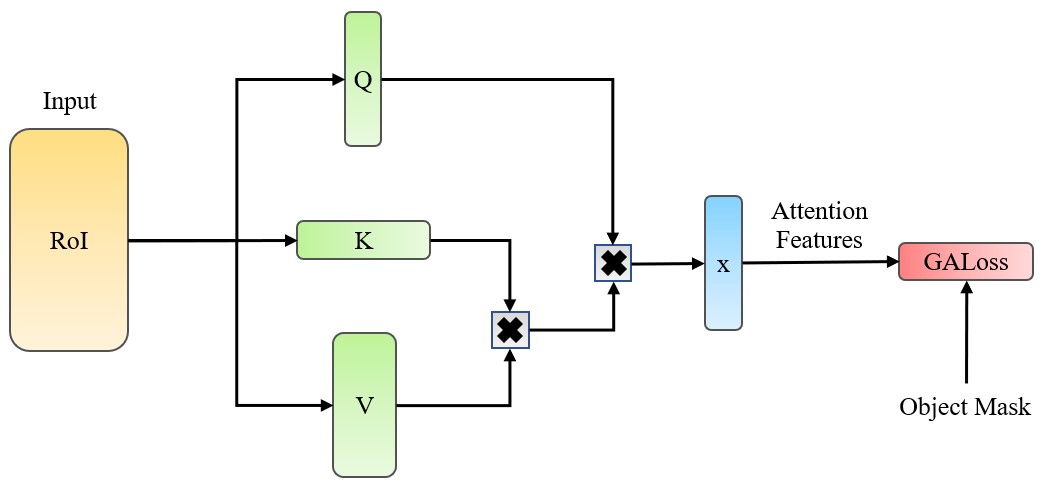}
\caption{Illustration of Guided-Attention Loss. Input RoI is transformed into three vectors Queries (Q), Keys (K), and Values (V), then processed by a self-attention network to obtain attention features (x) that are compared to object masks using Guided-Attention Loss.}
\label{fig:guidedattenjpg}
\end{figure}

\begin{algorithm}[t!]
\caption{Guided-Attention Loss ($GALoss$)}\label{alg:GALoss}
\begin{algorithmic}
    \Require RoI and Mask
    \Ensure $GALoss$ value
        \State $x \gets SA(RoI)$ \Comment{SA is self-attention}
        \State $y \gets Mask$
        \State $GALoss \gets -\frac{1}{N} \sum_{j=1}^{N} [y_j \log(x_j) + (1-y_j) \log(1-x_j)]$
\end{algorithmic}
\end{algorithm}

To make sure that we highlight the object in every RoI, we used a loss function that compares the attention features and object masks. These masks are obtained by converting the bounding box into instance segmentation of the object. We got this idea from \cite{Pang2019}, but instead of using CNN to produce attention features, we used self-attention to get the global context of objects from RoI's.

First, RoI's are processed by self-attention to obtain rich attention features, then we used binary cross-entropy to compare the similarity between the features and masks. Through training, the attention features will learn the object masks and start to focus on the foreground which will result into having more information than the input RoI. Using these features, it will improve the detection of small objects in aerial images and boost the performance on complex instances. The computation of GALoss is given in Algo. \ref{alg:GALoss} and shown in Fig. \ref{fig:guidedattenjpg}.

\subsection{Box-Points Loss ( \textbf{\textit{BPLoss}} )}

\begin{figure}[t!]
\centering
\includegraphics[width=0.9\linewidth]{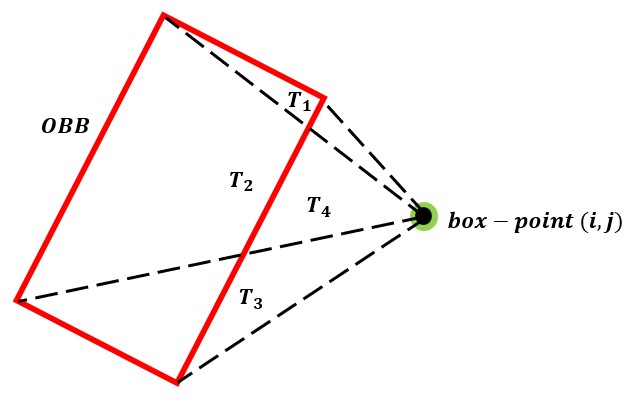}
\caption{General idea of Box-Points Loss. ${T_1, T_2, T_3, T_4}$ are the triangles formed when the edges of the OBB are connected with the box-point at ${(i,j)}$.}
\label{fig:OBBjpg}
\end{figure}

The \textit{BPLoss} is a function where we calculate the distance between an OBB and box-point located at \( \textbf{\textit{(i,j)}} \), illustrated in  Fig. \ref{fig:OBBjpg}. We compute the relative position of a box-point (inside or outside the box) as follows:

\begin{equation}
\label{eq:4.1}
\delta{(BP_{i,j}|OBB)} =
\begin{cases}
1, & \sum_{n=1}^{4} Area_{T_n} \le Area_{OBB} \\
0, & \text{otherwise}
\end{cases} 
\end{equation}

As given in (\ref{eq:4.1}), the \textit{BP} is the box-point and $Area_{OBB}$ is the area of the OBB. If the sum of the areas of the triangles is less than or equal to the $Area_{OBB}$, this means the \textit{BP} is inside the OBB, otherwise it is outside. Equation (\ref{eq:4.1}) is a non-differentiable function, meaning we cannot have useful gradients during training, so we designed a kernel function that can approximate (\ref{eq:4.1}), given by:

\begin{equation}
\label{eq:4.2}
\delta{(BP_{i,j}|OBB)} = \frac{2}{1+e^{k \frac{\sum_{n=1}^{4} Area_{T_n} - Area_{OBB}}{Area_{OBB}}}}
\end{equation}

Finally, to compute for the $BPLoss$, we subtract one by the sum of contributions of each kernel averaged by the total number of points. This is given by (\ref{eq:4.3}).

\begin{equation}
\label{eq:4.3}
BPLoss = 1 - \frac{\sum_{n=1}^{N} \delta{(BP_{n}|OBB)}}{N}
\end{equation}

The calculation of BPLoss is similar to PIoU Loss \cite{Chen2020}. In PIoU Loss, the distance of pixels and OBB center is computed. They used OBB of both the target and prediction while in BPLoss we only used the target OBB and did not have to convert the coordinates into bounding box center format since we only need the vertices. Furthermore, our distance calculation between the box-points and OBB is done through the difference of areas which is different from the PIoU Loss that used the euclidean distance of pixels.

\subsection{Evaluation}

The parameter we used to compute the detection score on small oriented objects is the mean average precision (mAP). To calculate the mAP, we need to know the area of intersection over the area of union (also called as IoU) between the ground-truth and prediction boxes. We can set different IoU thresholds to get the true positives (TP) and false positives (FP) of our predictions. If the IoU is greater than the threshold, the prediction is TP, otherwise, it is FP. With TP and FP, we can calculate the precision score, which is the number of true positives (TP) over the sum of all positive predictions (TP + FP) and its average is the weighted mean at every threshold. Finally, mAP is the average precision (AP) of each class (i) averaged over the total number of classes (N) given by:

\begin{equation}
\label{eq:mAP}
mAP = \frac{1}{N}\sum_{i=1}^{N} AP_i
\end{equation}

We used PASCAL VOC 2007 (VOC07) \cite{pascalvoc2007} and 2012 (VOC12) \cite{pascalvoc2012} mAP evaluation metrics in this paper. By default, we used IoU=0.5 and VOC07 in evaluating our model but we also employ (VOC12) and other IoU thresholds accepted as standards like IoU=0.7 and IoU=0.5:0.95.

\section{Experiments}

To evaluate our method, we used two datasets: DOTA-v1.5 \cite{Xia2018} for standard aerial images with small object instances and HRSC2016 \cite{icpram17} for maritime-related images.

\begin{table*}[t!]
\centering
\resizebox{.95\linewidth}{!}{\small
\begin{tabular}{|c|c|c|c|c|c|c|c|c|c|c|c|c|c|c|c|c|c|}
\hline
\multicolumn{18}{|c|}{Single Stage} \\
\hline
    Method & PL & BD & BR & GTF & SV & LV & SH & TC & BC & ST & SBF & RA & HA & SP & HC & CC & mAP\textsubscript{50} \\
\hline
    RetinaNet\textsuperscript{$\alpha$} \cite{Lin2020} & 0.753 & 0.754 & 0.318 & 0.619 & 0.321 & 0.692 & 0.790 & 0.896 & 0.716 & 0.588 & 0.430 & 0.661 & 0.501 & 0.603 & 0.390 & 1.5e-5 & 0.5649\\
    FCOS\textsuperscript{$\alpha$} \cite{tian2019fcos} & 0.786 & 0.725 & 0.443 & 0.595 & 0.562 & 0.640 & 0.780 & 0.894 & 0.714 & 0.733 & 0.495 & 0.664 & 0.557 & 0.632 & 0.447 & 0.094 & 0.6104\\
    RSDet++\textsuperscript{$\alpha$} \cite{Qian2021RSDetPM} & 0.793 & 0.740 & 0.449 & 0.609 & 0.564 & 0.605 & 0.780 & 0.894 & 0.708 & 0.735 & 0.512 & 0.683 & 0.562 & 0.682 & 0.522 & 0.102 & 0.6218\\
    DCL\textsuperscript{$\alpha$} \cite{yang2020dense} & 0.803 & 0.743 & 0.442 & 0.620 & 0.502 & 0.721 & 0.788 & 0.892 & 0.740 & 0.670 & 0.454 & 0.690 & 0.561 & 0.640 & 0.551 & 0.092 & 0.6198\\
    GWD\textsuperscript{$\alpha$} \cite{Yang2021c} & 0.802 & 0.742 & 0.465 & 0.638 & 0.566 & 0.736 & 0.823 & 0.899 & 0.742 & 0.704 & 0.475 & 0.688 & 0.598 & 0.650 & 0.46 & 0.113 & 0.6322\\
    BCD\textsuperscript{$\alpha$} \cite{https://doi.org/10.48550/arxiv.2209.10839} & 0.802 & 0.737 & 0.397 & 0.650 & 0.568 & 0.745 & 0.866 & 0.896 & 0.751 & 0.663 & 0.487 & 0.647 & 0.642 & 0.643 & 0.524 & 0.138 & 0.6353\\
    KLD\textsuperscript{$\alpha$} \cite{Yang2021a} & 0.802 & 0.727 & 0.473 & 0.601 & 0.632 & 0.751 & 0.860 & 0.895 & 0.735 & 0.729 & 0.503 & 0.662 & 0.645 & 0.691 & 0.578 & 0.137 & 0.6517\\
    KFIoU\textsuperscript{$\alpha$} \cite{https://doi.org/10.48550/arxiv.2201.12558} & 0.801 & 0.775 & 0.470 & 0.669 & 0.568 & 0.748 & 0.843 & 0.908 & 0.767 & 0.670 & 0.470 & 0.703 & 0.573 & 0.663 & 0.572 & 0.142 & 0.6469\\
    R\textsuperscript{2}CNN\textsuperscript{$\alpha$} \cite{Jiang2017R2CNNRR} & 0.803 & 0.787 & 0.479 & 0.625 & 0.656 & 0.713 & 0.863 & 0.897 & 0.762 & 0.762 & 0.497 & 0.675 & 0.634 & 0.731 & 0.584 & 0.156 & 0.6644\\
\hline
\multicolumn{18}{|c|}{Multi Stage} \\
\hline
    MR\textsuperscript{$\beta$} \cite{he2017} & 0.7684 & 0.7351 & 0.4990 & 0.5780 & 0.5131 & 0.7134 & 0.7975 & 0.9046 & 0.7421 & 0.6607 & 0.4621 & 0.7061 & 0.6307 & 0.6446 & 0.5781 & 0.0942 & 0.6267\\
    CMR\textsuperscript{$\beta$} \cite{Cai2018CascadeRD} & 0.6777 & 0.7462 & 0.5109 & 0.6344 & 0.5164 & 0.7290 & 0.7999 & 0.9035 & 0.7400 & 0.6750 & 0.4954 & 0.7285 & 0.6419 & 0.6488 & 0.5587 & 0.0302 & 0.6341\\
    HTC\textsuperscript{$\beta$} \cite{chen2019hybrid} & 0.7780 & 0.7367 & 0.5140 & 0.6399 & 0.5154 & 0.7331 & 0.8031 & 0.9048 & 0.7512 & 0.6734 & 0.4851 & 0.7063 & 0.6484 & 0.6448 & 0.5587 & 0.0515 & 0.6340\\
    FR\textsuperscript{$\beta$} \cite{Ren2017} & 0.7189 & 0.7447 & 0.4445 & 0.5987 & 0.5128 & 0.6880 & 0.7937 & 0.9078 & 0.7738 & 0.6750 & 0.4775 & 0.6972 & 0.6122 & 0.6528 & 0.6047 & 0.0154 & 0.6200\\
    RT\textsuperscript{$\beta$} \cite{Ding2019} & 0.7192 & 0.7607 & 0.5187 & 0.6924 & 0.5205 & 0.7518 & 0.8072 & 0.9053 & 0.7858 & 0.6826 & 0.4918 & 0.7174 & 0.6751 & 0.6553 & 0.6216 & 0.0999 & 0.6503\\
    ORCNN \cite{Xie2021} & 0.8098 & 0.8500 & \textcolor{red}{0.5992} & 0.7960 & 0.6775 & \textcolor{red}{0.8206} & 0.8978 & \textcolor{red}{0.9088} & \textcolor{red}{0.7893} & 0.7791 & \textcolor{red}{0.7097} & 0.7617 & \textcolor{red}{0.8173} & \textcolor{red}{0.7664} & 0.7354 & \textcolor{red}{0.4709} & 0.7619 \\
    OURS & \textcolor{red}{0.8620} & \textcolor{red}{0.8563} & 0.5914 & \textcolor{red}{0.8015} & \textcolor{red}{0.6780} & 0.8180 & \textcolor{red}{0.8989} & 0.9080 & 0.7789 & \textcolor{red}{0.7843} & 0.6977 & \textcolor{red}{0.7618} & 0.8125 & 0.7654 & \textcolor{red}{0.7536} & 0.4234 & \textcolor{red}{\textbf{0.7620}}\\
\hline
\multicolumn{18}{c}{} \\
\hline
    ORCNN* \cite{Xie2021} & 0.8011 & 0.6736 & 0.4590 & 0.6765 & 0.5912 & 0.7428 & 0.8750 & \textcolor{blue}{0.9074} & 0.6927 & 0.7519 & 0.4814 & 0.6971 & 0.6854 & 0.6636 & 0.5956 &  \textcolor{blue}{0.3798} & 0.6672 \\
    OURS* & \textcolor{blue}{0.8534} & \textcolor{blue}{0.8051} & \textcolor{blue}{0.5473} & \textcolor{blue}{0.7489} & \textcolor{blue}{0.6567} & \textcolor{blue}{0.7994} & \textcolor{blue}{0.8864} & 0.9067 & \textcolor{blue}{0.7584} & \textcolor{blue}{0.7758} & \textcolor{blue}{0.6319} & \textcolor{blue}{0.7264} & \textcolor{blue}{0.7235} & \textcolor{blue}{0.7314} & \textcolor{blue}{0.6206} & 0.3041 & \textcolor{blue}{\textbf{0.7172}} \\
\hline
\end{tabular}}
\caption{Comparison of results on DOTA-v1.5 trainval/test and train/test* splits (ORCNN \cite{Xie2021} is the baseline). The colors red and blue indicate the highest value of trainval/test and train/test, respectively. The $\alpha$ and $\beta$ denote that results are obtained from AlphaRotate \cite{yang2021alpharotate} and AerialDetection \cite{ding2021object} libraries, respectively.}
\label{table:IoUThresholds}
\end{table*}

\begin{table}[t]
\centering
\resizebox{\linewidth}{!}
    {
        \normalsize{
            \begin{tabular}{|c|c|c|c|}
            \hline
            Method & mAP\textsubscript{50} & mAP\textsubscript{75} & mAP\textsubscript{50:95} \\
            \hline
            Baseline \cite{Xie2021} & 0.7619 & 0.5089 & 0.4795 \\
            Ours & 0.7620 \textcolor{red}{(+0.01\%)} & 0.5204 \textcolor{red}{(+2.24\%)} & 0.4824 \textcolor{red}{(+0.59\%)} \\
            \hline
            Baseline* \cite{Xie2021} & 0.6672 & 0.3886 & 0.3800 \\
            Ours* & 0.7172 \textcolor{red}{(+7.49\%)} & 0.4425 \textcolor{red}{(+13.87\%)} & 0.4260 \textcolor{red}{(+12.11\%)} \\
            \hline
            \end{tabular}
        }
    }
\caption{Comparison of results on DOTA-v1.5 trainval/test and train/test* splits with different IoU thresholds. The color red indicate the relative difference between the methods.}
\end{table}

\subsection{Datasets}

The Dataset for Object Detection in Aerial Images version-1.5 (DOTA-v1.5) \cite{Xia2018} is the largest dataset for object detection in aerial images with oriented bounding box annotations. It contains 2806 large-size images (1/2 train, 1/6 val, and 1/3 test splits) with 403,318 instances and 16 categories including Plane (PL), Baseball diamond (BD), Bridge (BR), Ground track field (GTF), Small vehicle (SV), Large vehicle (LV), Ship (SH), Tennis court (TC), Basketball court (BC), Storage tank (ST), Soccer-ball field (SBF), Roundabout (RA), Harbor (HA), Swimming pool (SP), Helicopter (HC), and Container Crane (CC). DOTA-v1.5 contains extremely small instances (less than 10 pixels) that vary greatly in scale, orientation, and aspect ratio, that increases the difficulty of object detection. 

The High-Resolution Ship Collections 2016 (HRSC2016) \cite{icpram17} is a maritime-related dataset that contains ships from the sea and inshore. It contains 1061 images ranging from 300×300 to 1500×900 pixels for which train, val, and test sets have 436, 181, and 444 images, respectively. We combined the train and val sets for training and the test set for testing.


\subsection{Implementation}

We used Quadro RTX 8000 for training the models and OBBDetection \cite{Xie2021}, a PyTorch library that contains different sets of oriented object detection models modified from MMdetection toolbox \cite{mmdetection} to automatically check the performance. We also based the comparison of results published in AerialDetection \cite{ding2021object} and AlphaRotate \cite{yang2021alpharotate} libraries.

For DOTA-v1.5, we cropped a series of 1024x1024 patches from the original images with a stride of 524 and resized the images into multiple scales, 0.5x, 1.0x, and 1.5x with random rotation from 0-90 degrees. We optimized the network training using SGD algorithm with momentum of 0.9 and weight decay of 0.0001. We used two dataset splits for training and evaluation, trainval/test and train/test. The former is trained for 36 epochs and has an initial learning rate of 0.005 with learning rate scheduling that is divided by 10 at epochs 24 and 33, while the latter is trained for 20 epochs with no learning rate scheduling.

For HRSC2016, we randomly rotated the objects during training from 0-90 degrees, resized the images into 1333x800, and trained for 180 epochs with R-50-FPN as backbone.

\begin{table}[t]
\centering
\resizebox{.95\linewidth}{!}%
    {
        \normalsize{
            \begin{tabular}{|c|c|c|c|}
            \hline
            Method & Backbone & mAP\textsubscript{50}(07) & mAP\textsubscript{50}(12) \\
            \hline
            PIoU \cite{Chen2020} & DLA-34 & 0.8920 & - \\
            R3Det \cite{Yang2021b} & R-101-FPN & 0.8926 & 0.9601 \\
            DAL \cite{ming2021dynamic} & R-101-FPN & 0.8977 & - \\
            S2ANet \cite{Han2021a} & R-101-FPN & 0.9017 & 0.9501 \\
            Rotated RPN \cite{Ma2018} & R-101 & 0.7908 & 0.8564 \\
            R2CNN \cite{Jiang2017R2CNNRR} & R-101 & 0.7307 & 0.7973 \\
            RoI Transformer \cite{Ding2019} & R-101-FPN & 0.8620 & - \\
            Gliding Vertex \cite{Xu2021} & R-101-FPN & 0.8820 & - \\
            Oriented R-CNN \cite{Xie2021} & R-50-FPN & 0.9040 & 0.9650 \\
            Ours & R-50-FPN & \textbf{0.9059} & \textbf{0.9789} \\
            \hline
            \end{tabular}
        }
    }
\caption{Comparison of results on HRSC2016.}
\label{table:HRSCresults}
\end{table}
\begin{figure*}[t!]
\centering
    \includegraphics[height=2.5cm, width=.192\linewidth]{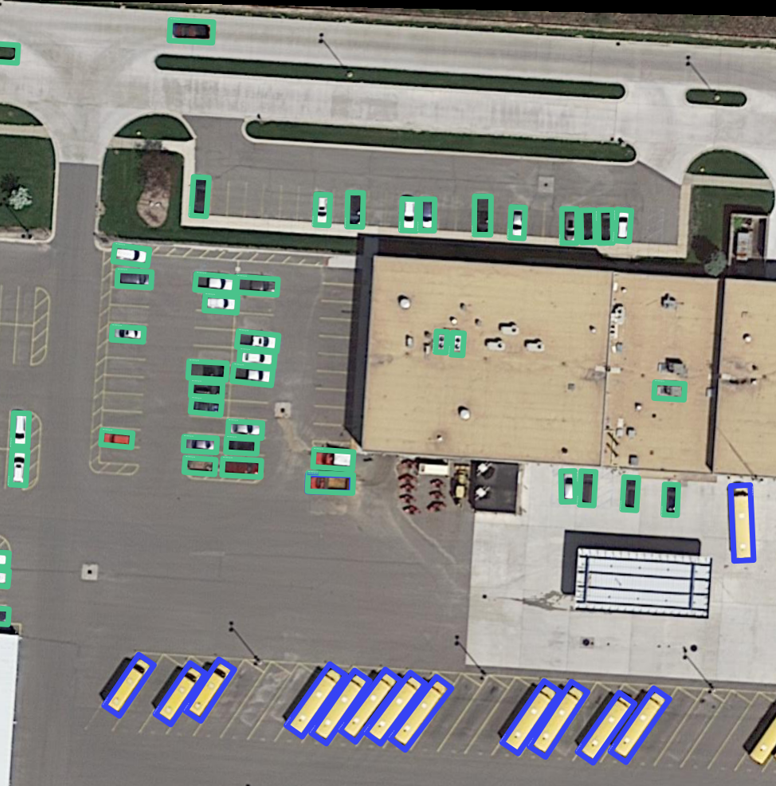}\hfill
    \includegraphics[height=2.5cm, width=.192\linewidth]{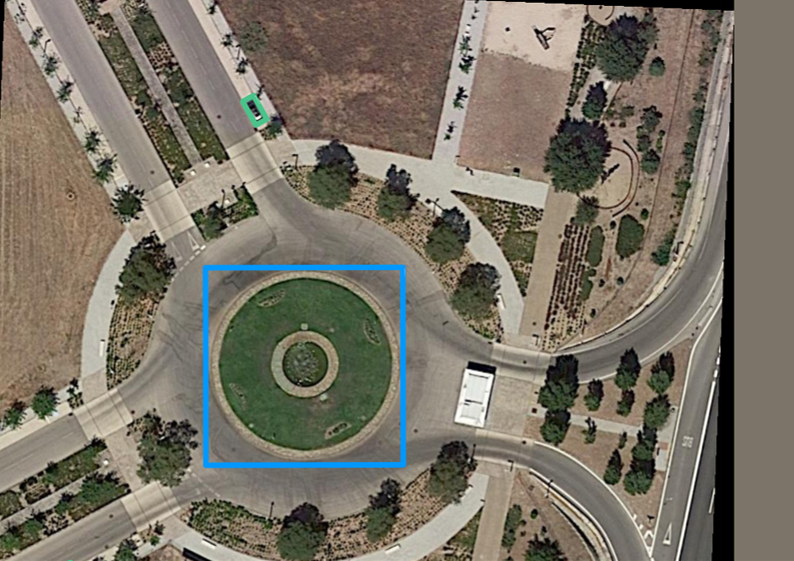}\hfill
    \includegraphics[height=2.5cm, width=.192\linewidth]{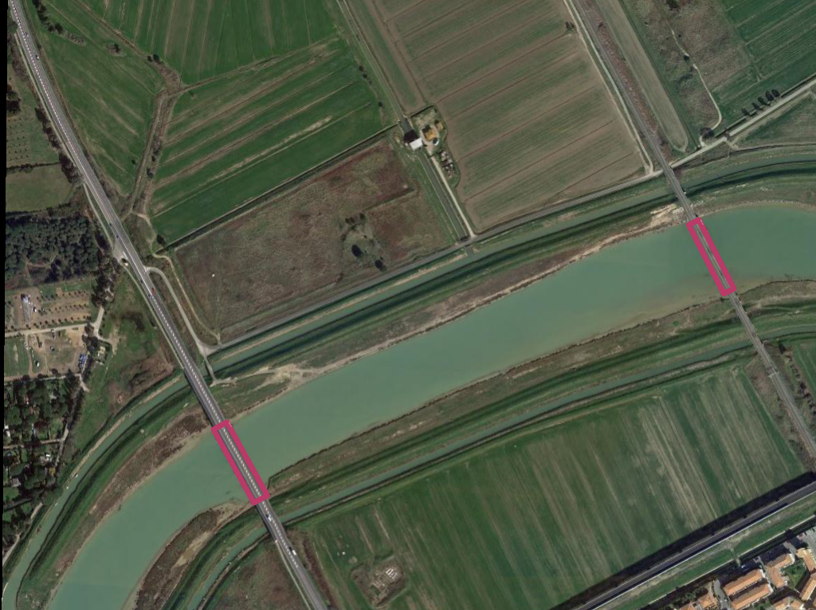}\hfill
    \includegraphics[height=2.5cm, width=.192\linewidth]{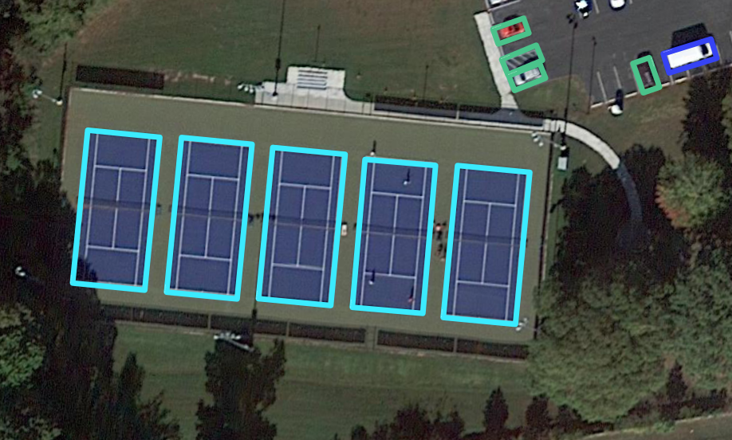}\hfill
    \includegraphics[height=2.5cm, width=.192\linewidth]{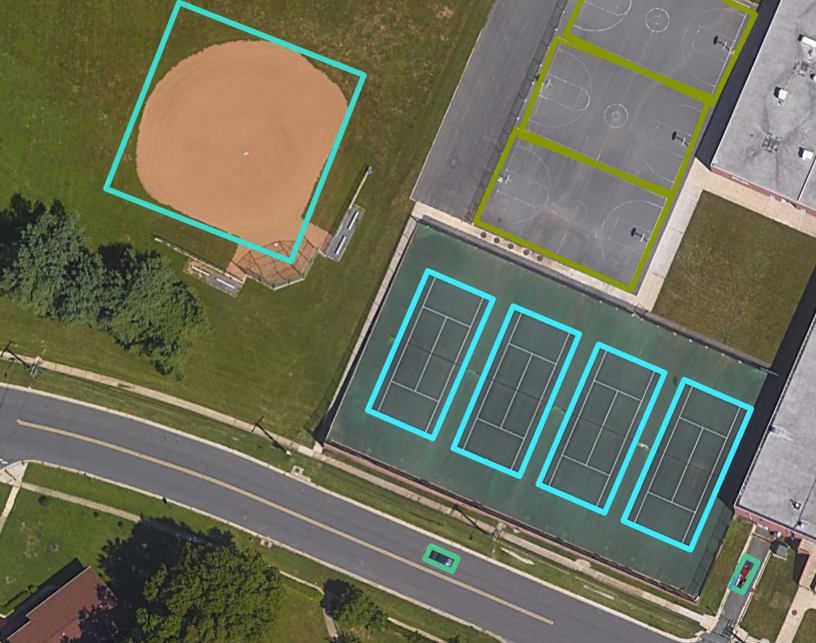}
    \\[\smallskipamount]
    \includegraphics[height=2.5cm, height=2cm,width=.192\linewidth]{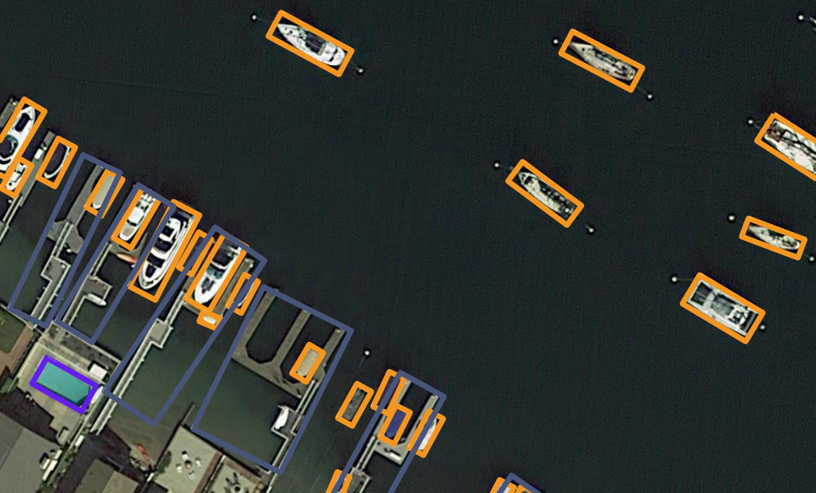}\hfill
    \includegraphics[height=2.5cm, height=2cm,width=.192\linewidth]{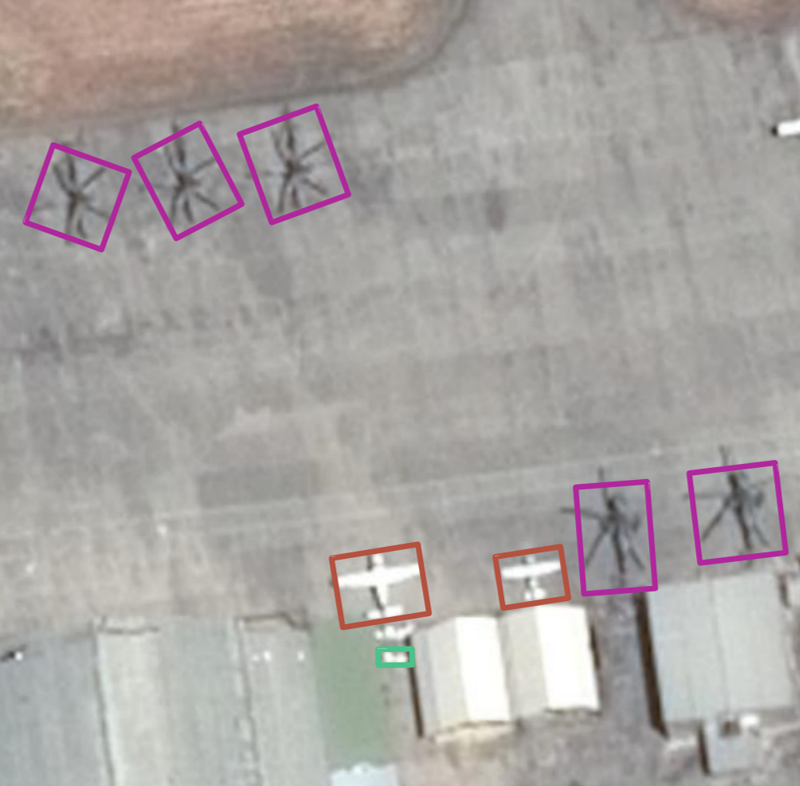}\hfill
    \includegraphics[height=2.5cm, height=2cm,width=.192\linewidth]{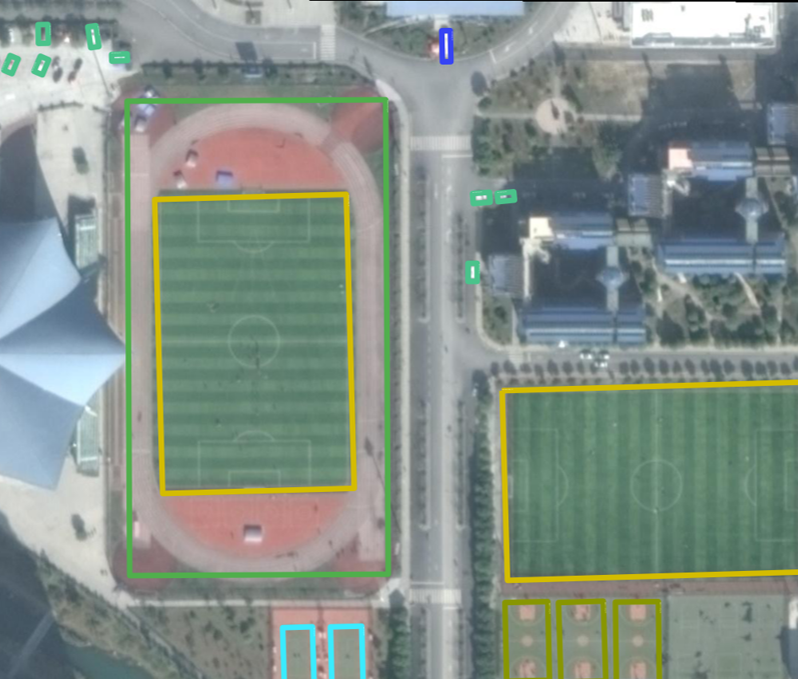}\hfill
    \includegraphics[height=2.5cm, height=2cm,width=.192\linewidth]{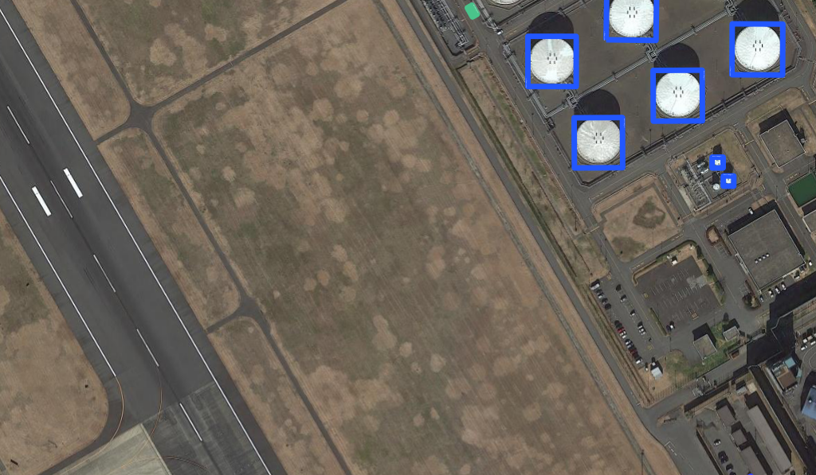}\hfill
    \includegraphics[height=2.5cm, height=2cm,width=.192\linewidth]{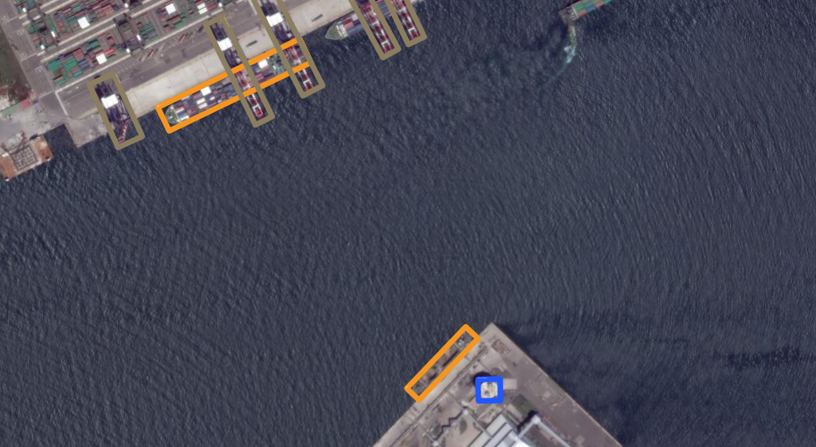}
    \\[\smallskipamount]
    \includegraphics[width=0.96\linewidth]{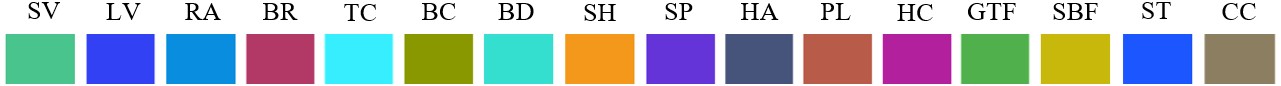}
\caption{Visualization of detection results on DOTA-v1.5 dataset.}
\label{fig:DOTAvisualization}
\end{figure*}

\begin{figure*}[t!]
\centering
    \includegraphics[height=3cm, width=.24\linewidth]{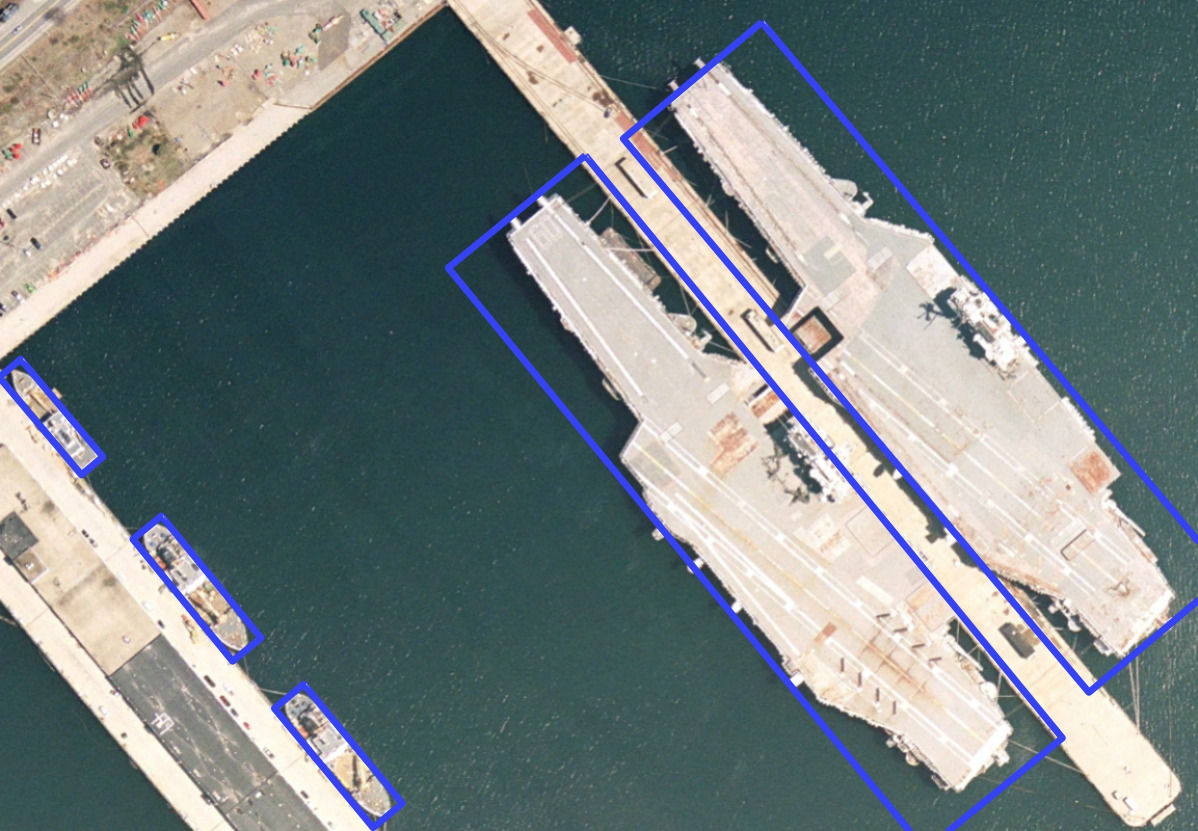}\hfill
    \includegraphics[height=3cm,width=.24\linewidth]{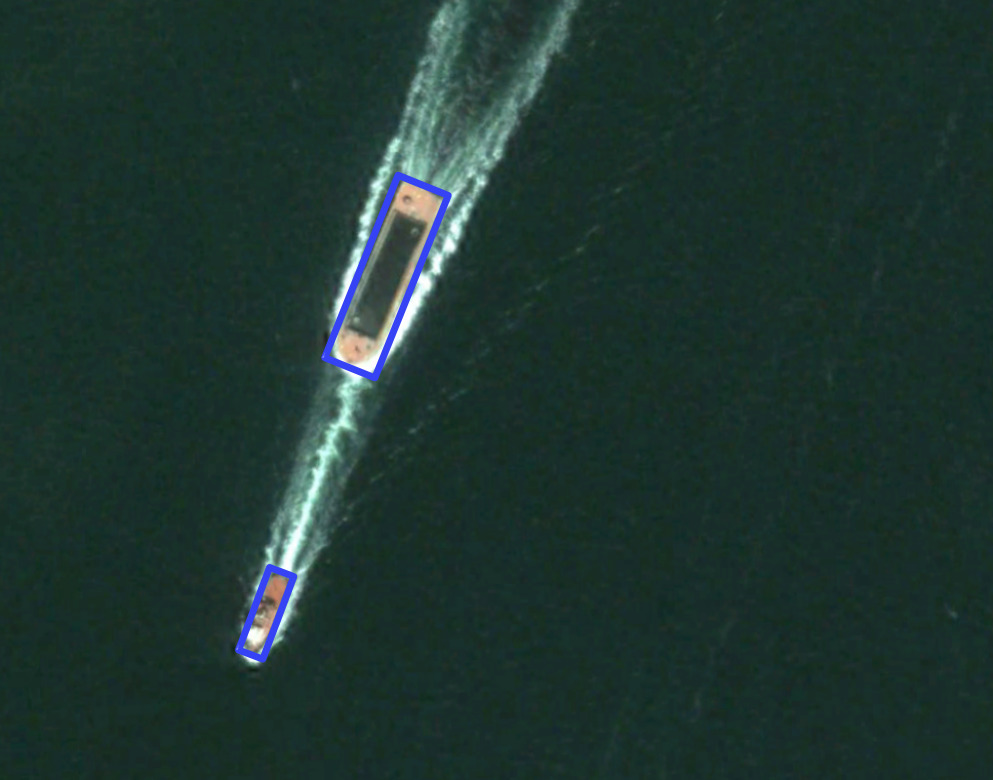}\hfill
    \includegraphics[height=3cm,width=.24\linewidth]{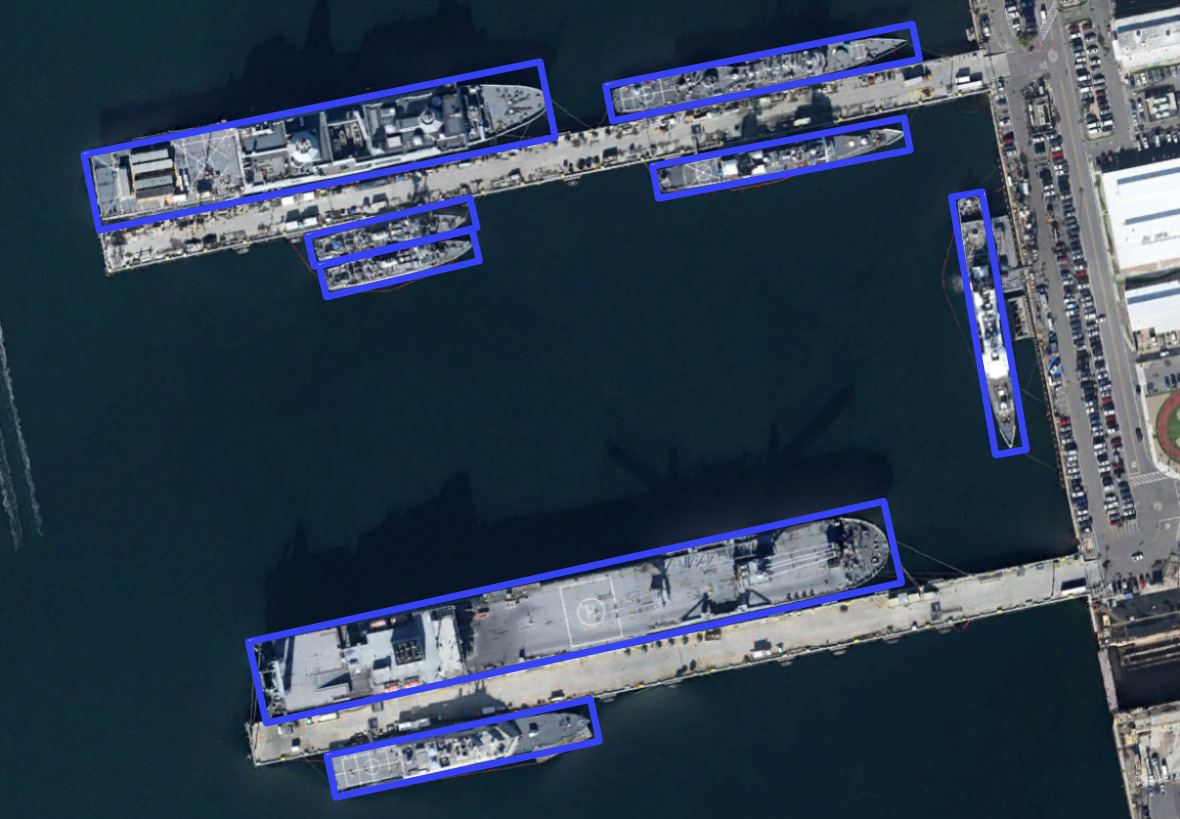}\hfill
    \includegraphics[height=3cm,width=.24\linewidth]{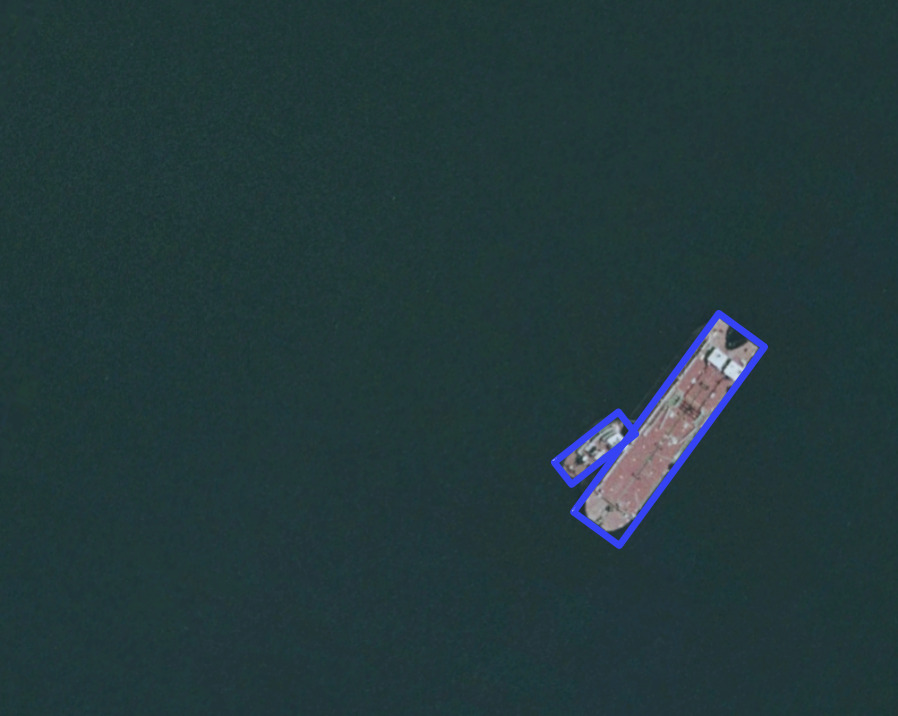}
\caption{Visualization of detection results on HRSC2016 dataset. Ships are either in the sea or inshore.}
\label{fig:HRSCvisualization}
\end{figure*}


\subsection{Comparison with other Methods}

\begin{figure}[t]
\centering
\includegraphics[height=5.7cm, width=0.9\linewidth]{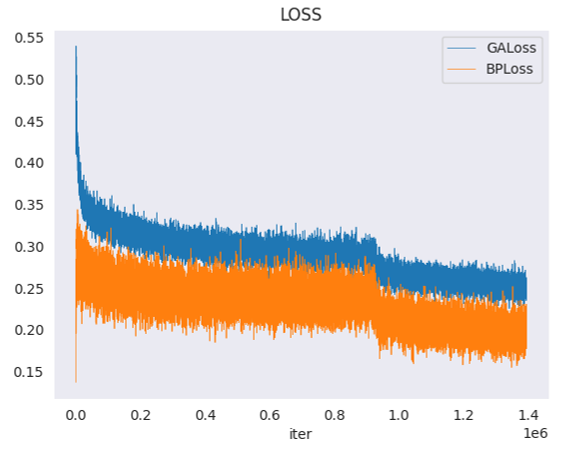}
\caption{Learning curves of GALoss and BPLoss.}
\label{fig:lossgraphs}
\end{figure}



\textbf{Results on DOTA-v1.5:} We compared our results with other methods. As shown in Table \ref{table:comparisonSOTA}, our method  has a marginal increase of mAP\textsubscript{50} over the baseline at trainval/test split, but obtained a large improvement of 7.5\% when using the train/test split. Furthermore, the classes with the smallest instances: small vehicles and ships, have increased by 11.01\% and 1.3\%, respectively. The reason why our method performed marginally at trainval/test and better in train/test is because of the distribution of data in validation set. The validation set contain images with more complex instances than the training set, hence the difference in performance. Moreover, the published results of related works only showed the mAP\textsubscript{50} on trainval/test, so we implemented the baseline using train/test and compared with our method. Finally, the baseline has the second highest performance in DOTA-v1.5 trainval/test that is why we chose it for comparison in train/test split.

In Table \ref{table:IoUThresholds}, we showed that our method achieved a very high performance increase versus the baseline when evaluated both on DOTA-v1.5 trainval/test and train/test splits across mAP\textsubscript{75} and mAP\textsubscript{50:95} evaluation metrics. We attributed the difference of performance of mAP\textsubscript{75} and mAP\textsubscript{50:95} to mAP\textsubscript{50} on the nature of small oriented aerial objects. These objects represents only a tiny pixel-area in an image so using IoU=0.5 is a coarse threshold and could miss objects with small instances. Thus, using finer thresholds of IoU=0.75 and IoU=0.5:0.95 are more appropriate in metrics. Figure \ref{fig:DOTAvisualization} shows sample detection results on DOTA-v1.5.

\textbf{Results on HRSC2016:} We also compared our method on this dataset to the baseline and other methods, shown in Table \ref{table:HRSCresults}. We used mAP\textsubscript{50} of PASCAL VOC 2007 and VOC 2012 metrics to compare the performance. As can be seen, our method achieved results in mAP\textsubscript{50}(07) and mAP\textsubscript{50}(12) that are better than the baseline. Visualization of results on HRSC is shown in Fig \ref{fig:HRSCvisualization}. 

\begin{table}[t]
\centering
\resizebox{.95\linewidth}{!}
    {
        \normalsize{
            \begin{tabular}{|l|c|c|}
            \hline
            \multicolumn{1}{|c|}{Method} & mAP\textsubscript{75} & mAP\textsubscript{50:95} \\
            \hline
            \textbf{Case 0}: Baseline \cite{Xie2021} & 0.5089 & 0.4795 \\
            \textbf{Case 1}: GALoss only & 0.5144 \textcolor{red}{(+1.08\%)} & 0.4826 \textcolor{red}{(+0.642\%)} \\
            \textbf{Case 2}: BPLoss only & 0.5171 \textcolor{red}{(+1.61\%)} & 0.4815 \textcolor{red}{(+0.415\%)} \\
            \textbf{Case 3}: GALoss and BPLoss & 0.5204 \textcolor{red}{(+2.26\%)} & 0.4824 \textcolor{red}{(+0.605\%)} \\
            \hline
            \end{tabular}
        }
    }
\caption{Ablation experiment conducted on DOTA-v1.5 trainval/test.}
\label{table:ablationtable}
\end{table}

\subsection{Loss Functions}
 To show that our loss functions learned during training, we plotted the learning curves over the number of iterations and showed that GALoss and BPLoss are decreasing, illustrated in Fig. \ref{fig:lossgraphs}. Although the plots are noisy, this is expected since the ground-truths used for loss calculation are coarse-level like the object masks and target OBB. Note that the figure is not a comparison of which loss function contributed more to our method, but rather a visualization of how both loss functions learned during training.


\subsection{Ablation Study}

To evaluate the effectiveness of each loss functions, we conducted ablation experiment on DOTA-v1.5 dataset and compared the performance on the baseline using mAP\textsubscript{75} and mAP\textsubscript{50:95} evaluation metrics, shown in Table \ref{table:ablationtable}. Cases 1 and 2 are evaluated separately and measured their performance. As can be seen in the table, both cases contributed to the overall performance without lagging behind the baseline. We can also notice that the performance of mAP\textsubscript{50:95} at Case 3 did not sum up when we add the results of Case 1 and Case 2. This is because mAP\textsubscript{50:95} is an average performance when IoU=0.5 to IoU=0.95 is calculated, IoU=0.5 is a coarse threshold not appropriate for small instances which affects the calculation of mAP\textsubscript{50:95}. This is also the reason why we did not include it in the ablation experiment. Finally, if we look at the relative increase of Case 1  and Case 2 then take the mean, we can see that Case 3 is above the average. This shows that our designed loss functions are individually effective in the overall architecture of the Attention-Points Network.

\section{Conclusion}

We developed the Attention-Points Network and designed loss functions: Guided-Attention Loss (GALoss) and Box-Points Loss (BPLoss) for small oriented objects in aerial images. Results showed that our method was able to achieve better results against the baseline and other architectures on a standard oriented aerial dataset with small object instances (DOTA-v1.5) and on a maritime-related dataset (HRSC2016). Ablation experiment and learning curves of loss functions, GALoss and BPLoss, are also presented to verify the effectiveness of our method.

\section*{Acknowledgements} 

We would like to thank the Department of Science and Technology - Science Education Institute (DOST-SEI) for funding our research and providing graduate student scholarships through the Space Science and Technology Proliferation through University Partnerships (STeP-UP) project. A big thanks also to Dr. Rowel Atienza and the technical support of the people at the Computer Networks Laboratory (CNL) of Electrical and Electronics Engineering Institute (EEEI), University of the Philippines Diliman for allowing us to use their GPU's, servers and IT resources.

{\small
\bibliographystyle{ieee_fullname}
\bibliography{egbib}
}

\end{document}